\documentclass[letterpaper, 10 pt, conference]{ieeeconf}  

\IEEEoverridecommandlockouts                              

\usepackage{times}
\usepackage{amsmath}
\usepackage{amssymb}
\usepackage{caption}
\usepackage{mathtools}
\usepackage{graphicx}
\usepackage{multirow}
\usepackage{makecell}
\usepackage{booktabs}
\usepackage{algorithm}
\usepackage{algorithmic}
\usepackage{url}
\usepackage{hyperref}

\def \MethodAcronym {ViTaMIn}

\title{\LARGE \bf \MethodAcronym: Learning Contact-Rich Tasks Through Robot-Free Visuo-Tactile Manipulation Interface}

\author{Fangchen Liu$^{*,2}$, Chuanyu Li$^{*,1}$, Yihua Qin$^{1}$, Jing Xu$^{1}$, Pieter Abbeel$^{2}$, Rui Chen$^{\dagger,1}$ \\
${}^{1}$Tsinghua University, ${}^{2}$University of California, Berkeley\\
{\small $^{*}$ Equal contribution, 
$^{\dagger}$ Corresponding author} %
\\
\url{https://chuanyune.github.io/ViTaMIn_page}}

\begin{document}

\let\oldtwocolumn\twocolumn
\renewcommand\twocolumn[1][]{%
    \oldtwocolumn[{#1}{
    \begin{center}
           \includegraphics[width=\textwidth]{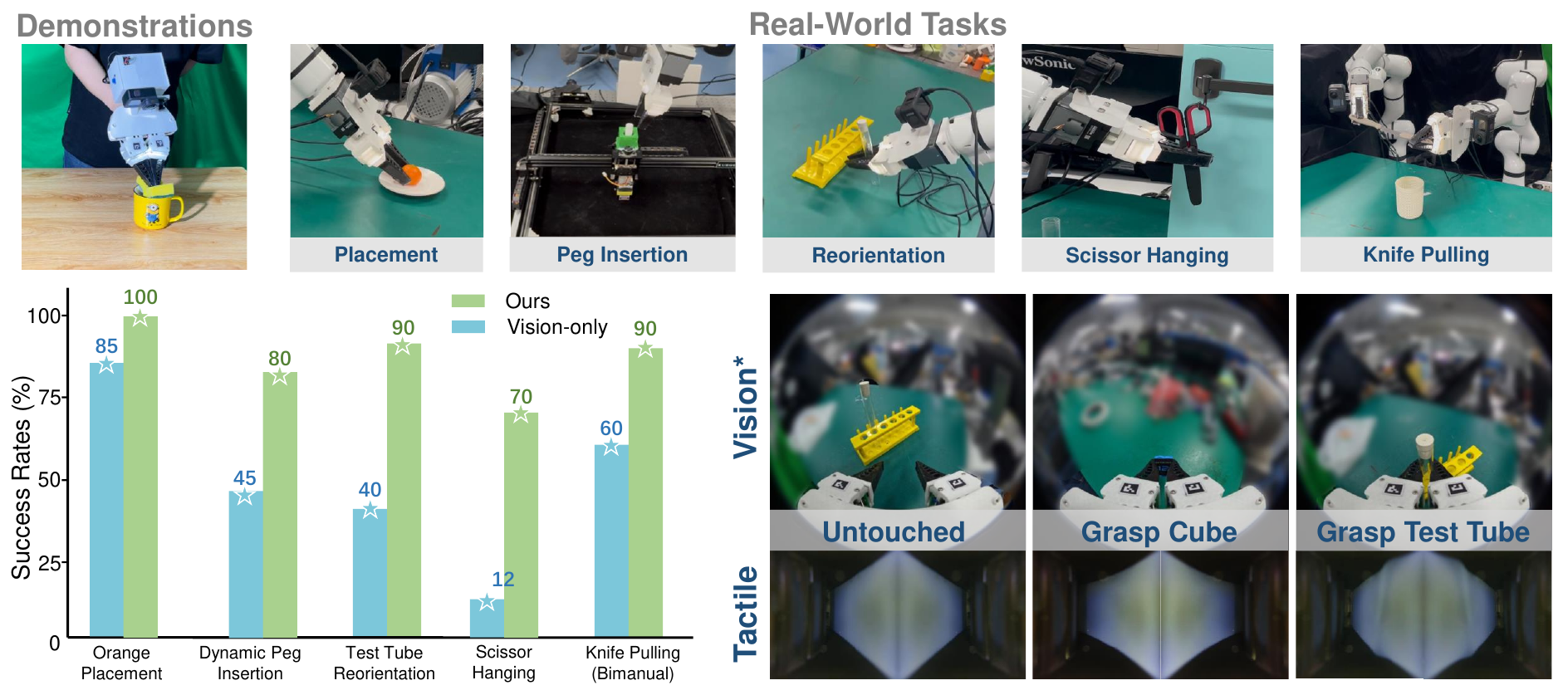}
           \captionof{figure}{\MethodAcronym{} overview. Our system comprises a portable data collection device that integrates visual and tactile sensing, a multimodal representation learning framework for fusing visual and tactile information, and demonstrations of various contact-rich manipulation tasks. This system facilitates efficient collection of manipulation data without requiring complex robot setups. (*Backgrounds in the images are blurred.)}
           \label{fig:teaser}
    \end{center}
    }]
}

\maketitle
\thispagestyle{empty}
\pagestyle{empty}

\begin{abstract}
Tactile information plays a crucial role for humans and robots to interact effectively with their environment, particularly for tasks requiring the understanding of contact properties. Solving such dexterous manipulation tasks often relies on imitation learning from demonstration datasets, which are typically collected via teleoperation systems and often demand substantial time and effort. To address these challenges, we present \MethodAcronym, an embodiment-free manipulation interface that integrates visual and tactile sensing into a hand-held gripper, enabling multi-modality data collection without the need for teleoperation. Our design employs a compliant Fin Ray gripper with tactile sensing, allowing operators to perceive force feedback during manipulation for more intuitive operation. Additionally, we propose a multi-modal representation learning strategy to obtain pre-trained tactile representations, improving data efficiency and policy robustness. Experiments on 5 contact-rich manipulation tasks demonstrate that our system is more scalable, efficient, and effective than baseline methods.

\end{abstract}

\section{Introduction}
Humans rely on both visual and tactile modalities to perform a diverse range of manipulation tasks in daily life. For instance, when inserting a plug into a socket or tightening a screw, vision helps with identifying and aligning components, while tactile signals enable precise force control during contact. This seamless integration of vision and touch enhances human dexterity, particularly in tasks that require contact-rich control, handling visual occlusions, or performing in-hand manipulations. 

Recent progress in learning from demonstrations \cite{levine2016end, brohan2022rt, brohan2023rt, chi2023diffusion} has shown significant potential for advancing general-purpose robots, enabling them to efficiently acquire complex skills from human demonstrations. Consequently, developing systems to collect high-quality demonstration data has been a recent key focus. Prior works have explored real-world data collection methods, including joint-mapped devices and exoskeletons \cite{aldaco2024aloha, fu2024mobile, zhao2023learning, fang2024airexo}, and vision-based teleoperation frameworks \cite{cheng2024open, qin2023anyteleop}. Nevertheless, these techniques require real-time teleoperation of a physical robot during data collection, which constrains efficiency and flexibility. In contrast, portable devices \cite{10341661, doshi2023hand, shafiullah2023bringing, chi2024universal} present a more scalable and cost-effective alternative to collect demonstration without teleoperation. Moreover, they can be seamlessly integrated into various embodiments, providing a more flexible data collection approach. However, these portable devices primarily focus on capturing vision-only demonstration data, limiting their usage for contact-rich and dexterous manipulation tasks where tactile feedback plays a crucial role.

In this work, we aim to address both the challenge of efficient data collection and the need for learning more dexterous tasks using visuo-tactile demonstrations. To this end, we introduce \MethodAcronym, a novel and effective visuo-tactile manipulation interface designed to capture high-quality demonstrations with enhanced efficiency and flexibility. Unlike conventional approaches that rely on rigid tactile sensors, \MethodAcronym{} leverages an omnidirectional compliant Fin Ray gripper with customized tactile sensing, which can detect contact from all directions as an expressive tactile signal for robot manipulation. We integrate the tactile-aware Fin Ray gripper~\cite{liang2025alltact} with UMI~\cite{chi2024universal}, enhancing the collected data with rich multimodal information and improving policy learning performance while maintaining the core advantages of portable devices. Additionally, our system enables operators to perceive force feedback during manipulation, facilitating more intuitive and seamless operation.

Pre-trained visual representations have shown improved performance in robotic manipulation \cite{nair2022r3m, ma2022vip, xiao2022masked, radosavovic2023real, majumdar2023we}, benefiting from large-scale visual pre-training. To fully leverage the visuo-tactile datasets collected with \MethodAcronym{}, we adopt a multimodal representation learning strategy to pre-train tactile representations, enhancing the robustness and generalizability of our sensor-based policies. Our pre-training objective integrates masked autoencoding~\cite{he2022masked} and contrastive learning for multimodal alignment~\cite{radford2021learning}, where future image observations are aligned with masked current images and tactile signals. Through extensive experiments on five challenging contact-rich manipulation tasks, our visuo-tactile policy, enhanced by multimodal pre-training, exhibits superior data and training efficiency while demonstrating strong generalization across diverse objects and environmental conditions.

In conclusion, our contributions are:
\begin{itemize}
    \item  \MethodAcronym{} provides a portable and scalable visuo-tactile data collection system.
    
    \item \MethodAcronym{} proposes an effective multimodal representation learning strategy, which significantly improves the data efficiency, robustness and generalization capabilities.

    \item \MethodAcronym{} achieves superior performance over vision-only baselines across five manipulation tasks by leveraging visuo-tactile demonstrations.
    \vspace{-2mm}
\end{itemize} 

\section{Related Work}
\subsection{Visuo-Tactile Manipulation}
Tactile sensing is essential for robotic manipulation as it provides signals about physical contact in addition to visual observation. Early works~\cite{hosoda1996adaptive, nakagaki1997study, miller1999integration} use RGB cameras and force/torque sensors to infer contact status for making decisions. However, the information from force/torque sensors is low-dimensional and insufficient for more dexterous manipulation tasks. 

More recently, vision-based tactile sensors have gained attention for their ability to capture high-resolution contact information~\cite{qi2023general, li2023visual, han2024learning}. Despite these advances, the rigid design of these sensors restricts the compliance of the end effector, where alternative approaches like uncalibrated tactile skins~\cite{bhirangi2024anyskin} and plug-and-play sensing systems~\cite{pattabiraman2024learning} have improved adaptability and flexibility. In our work, we use a Fin-Ray-shaped compliant and all-directional tactile sensor, which can detect contacts from all directions and also support safe and robust contact-rich manipulation.

\subsection{Data Collection System for Robot Manipulation}
Recent advancements in learning from demonstrations \cite{levine2016end, brohan2022rt, brohan2023rt, chi2023diffusion} have shown promising results in developing general-purpose robots. Therefore, efficiently collecting high-quality demonstrations has become a key research focus. 

Recently works have focused on efficient real-world data collection systems, such as devices or exoskeletons with joint-mapping~\cite{aldaco2024aloha, fu2024mobile, zhao2023learning},  exoskeletons~\cite{fang2024airexo}, or vision-based systems~\cite{cheng2024open, qin2023anyteleop}. However, these approaches require a physical robot during data collection, which limits efficiency and flexibility. In contrast, portable devices \cite{10341661, doshi2023hand, shafiullah2023bringing, chi2024universal, liu2024fastumi, liu2024maniwav} offer several advantages: they are low-cost, flexible, and do not depend on a specific physical robot. Additionally, they can be seamlessly integrated into various embodiments and provide a more user-friendly experience for data collection. We extend the UMI data collection system \cite{chi2024universal} by integrating tactile sensing, which enriches the demonstrations with multimodal information, improving policy learning performance while preserving the key benefits of portable devices. 

\subsection{Multimodal Pre-training for Robotics}
Pre-trained visual representations have shown improved performance and generalization in robotic manipulation~\cite{nair2022r3m, ma2022vip, xiao2022masked, radosavovic2023real, majumdar2023we} with self-supervised learning techniques~\cite{he2022masked, radford2021learning}. This can be extended to multimodal representation learning \cite{sferrazza2024power, xu2024unit, zhang2022fusing} by integrating visual, tactile, and proprioceptive modalities, allowing robots to perceive object properties beyond visual appearance.

Aligning heterogeneous sensory modalities is a key challenge in multimodal learning, as different sensors have varying data structures, sampling rates, and noise characteristics \cite{nagabandi2020deep}. Inspired by CLIP \cite{radford2021learning}, researchers have developed contrastive learning techniques to align tactile and visual representations for manipulation tasks \cite{fu2024a, yang2024unitouch}. 

Our work extends these efforts by introducing masked contrastive pre-training, where the tactile encoder learns to reconstruct future occluded visual information, further enhancing multimodal understanding.
\section{Visuo-Tactile Manipulation Interface}

\subsection{System Overview}
\begin{figure*}[!t]
\centering
\includegraphics[width=\textwidth]{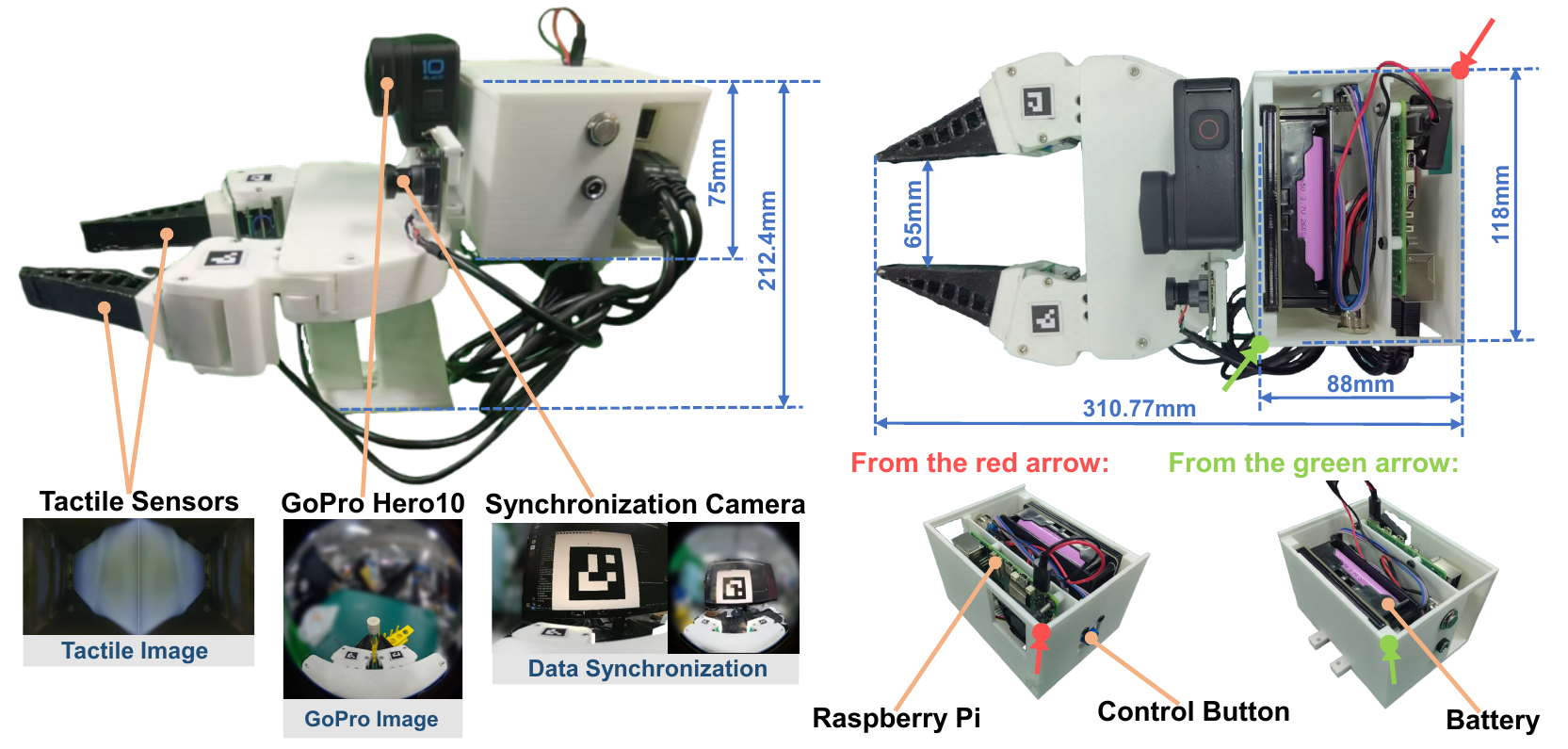}
\caption{
\MethodAcronym's hardware system overview. The handheld device integrates a GoPro camera, two tactile sensors and a synchronization camera to align visual and tactile information. During data collection, the two tactile sensors and the synchronization camera are connected to the Raspberry Pi in the backbox. The total weight of the gripper is approximately 1960g.
Left: Side view of the \MethodAcronym{} system.
Right: Top view of the \MethodAcronym{} system with the backbox cover removed.
}
\label{fig:system_overview}
\vspace{-5mm}
\end{figure*}

We design a handheld gripper to collect visuo-tactile demonstrations without requiring teleoperation on physical robots. Our gripper design is illustrated in Figure~\ref{fig:system_overview}. The gripper consists of an RGB fisheye wrist camera (GoPro 10) for image observation, two AllTact finger~\cite{liang2025alltact}, a synchronization camera for observation temporal alignment, and a Raspberry Pi 5 with a battery for data recording.

\textbf{Image Observation} To capture comprehensive visual information, we employ a GoPro 10 camera with a 155${}^\circ$ field-of-view (FoV) fisheye lens. The camera operates at 60 FPS with a resolution of 2704×2028 pixels and is mounted at the end-effector of our \MethodAcronym{} to ensure consistent visual coverage of the manipulation workspace during demonstration collection and policy deployment.

\textbf{Tactile Observation} 
In UMI~\cite{chi2024universal}, two TPU-printed Fin Ray grippers are used to provide compliance and enhance grasping stability. However, these grippers lack tactile sensing capabilities. In our \MethodAcronym{}, we employ AllTact~\cite{liang2025alltact}, a  compliant Fin Ray gripper with omnidirectional tactile sensing ability.
During manipulation, the embedded camera in AllTact captures both the global deformation of the entire finger and the local deformation of the contact surface as a single image. The tactile sensor operates at 30 FPS with a resolution of 640×480 pixels.

\textbf{Other Observations}
To enhance the robustness and accuracy of SLAM, we utilize the IMU data provided by the GoPro, which is synchronized with the visual observations. Gripper width is also critical for precise manipulation. Following UMI~\cite{chi2024universal}, we attach two ArUco markers to the gripper’s fingers and compute the gripper width from the visual observations.

\subsection{Data Processing}

\textbf{Sensor Synchronization} To synchronize the tactile sensors and GoPro camera, we use an additional low-cost camera which is connected to the Raspberry Pi and is naturally synchronized with the tactile sensors.
Before data collection, both the GoPro and the synchronization camera simultaneously capture a sequence of ArUco markers displayed on a computer screen. The ArUco IDs are detected in both video streams, and when an identical ID appears in both, the corresponding timestamps are used for synchronization.
Since the framerates of the GoPro and the synchronization camera are 60Hz and 30Hz respectively, the temporal alignment error is below $1/60 + 1/30 = 0.05$ seconds, which is sufficient for our tasks. Once the two videos are synchronized, they are cropped by the starting and ending signals triggered by the control button.

\textbf{Data Collection and Filtering} We adopt a similar data collection pipeline to UMI~\cite{chi2024universal}. We also utilize Simultaneous Localization and Mapping (SLAM) to capture the end-effector trajectories. While SLAM may fail in low-texture environments, it achieves a success rate of approximately $80\%$ in our tasks, allowing the majority of collected data to be used for imitation learning.

\section{Visuo-Tactile Policy Learning}
\subsection{Visuo-Tactile Representation Learning}
UMI uses a pre-trained CLIP~\cite{radford2021learning} encoder to extract visual representations. However, the tactile images in \MethodAcronym{} are very different from the CLIP's training distribution, which can lead to suboptimal representation. To tackle this, we pre-train an effective tactile encoder using the collected action-free datasets, which doesn't rely on the SLAM success. 

\begin{figure}[htbp!]
\centering
\includegraphics[width=1.0\columnwidth]{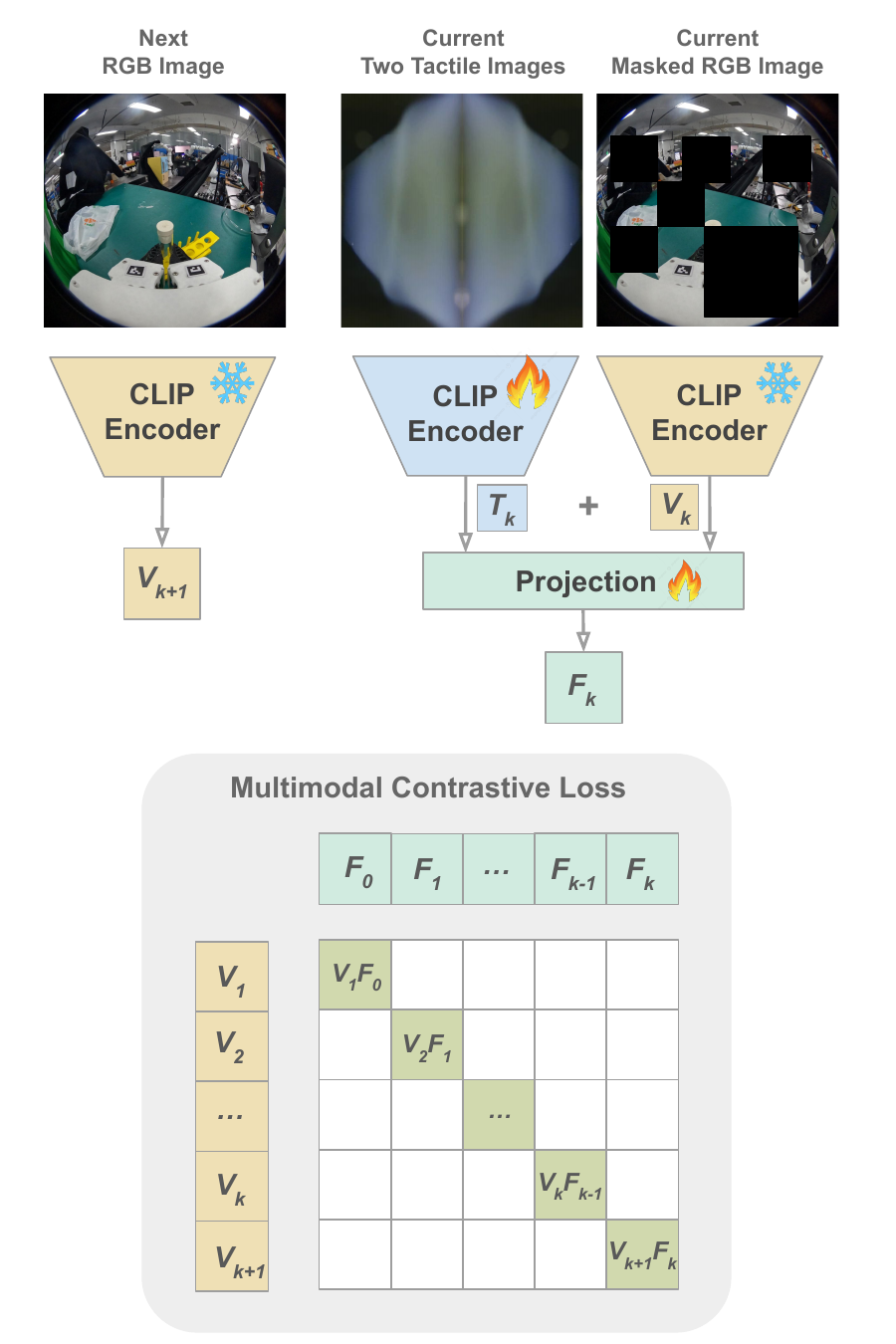}
\caption{The illustration of the multimodal contrastive representation pre-training phase. The tactile encoder is trained to capture complementary information to predict the missing content for the future image.}
\label{fig:pretraining}
\vspace{-3mm}
\end{figure}

Taking the tactile image in Figure~\ref{fig:pretraining} as an example, we want the encoder to capture the essential contact properties, such as the object's in-hand pose and gripper's deformation. These signals are complementary information from pixel observations, and are crucial for making future decisions. 

To achieve this, we employ a multimodal contrastive learning approach as illustrated in Figure~\ref{fig:pretraining}. Given the current masked image $\Tilde{I}_{V}^{k}$ and current full tactile observation $I_{T}^{k}$ of step $k$, we want the combination of $\Tilde{I}_{V}^{k}$ and $I_{T}^{k}$ align with the future full image observation $I_{V}^{k+1}$ in the CLIP embedding space. The intuition behind this is to make the tactile encoder focus on the contact information to predict future images based on the current corrupted image. 

To ensure stable training, we freeze the image CLIP encoder $\phi_V(\cdot)$ but only fine-tune the tactile encoder $\phi_T(\cdot)$. We first obtain the tactile embedding $T_k$ from $\phi_T(I_{T}^{k})$, and $V_k$ from $\phi_V(\Tilde{I}_{V}^{k})$. These embeddings are concatenated and passed through a fully connected projection layer, mapping them back to the original 512-dimensional CLIP embedding space as a fused feature $F_k$. Finally, we train the tactile encoder using the standard CLIP loss on $F_k$ and $V_{k+1}$: 
\begin{equation}
\mathcal{L}_{\text{CLIP}} = \frac{1}{2} \left( \mathcal{L}_{\text{f-v}} + \mathcal{L}_{\text{v-f}} \right)
\end{equation}
where
\begin{equation}
\mathcal{L}_{\text{v-f}} = -\frac{1}{N} \sum_{i=1}^{N} \log \frac{\exp(\text{cos}(V_{i+1}, F_i) / \tau)}{\sum_{j=1}^{N} \exp(\text{cos}(V_{i+1}, F_j) / \tau)}
\end{equation}

\begin{equation}
\mathcal{L}_{\text{f-v}} = -\frac{1}{N} \sum_{i=1}^{N} \log \frac{\exp(\text{cos}(F_i, V_{i+1}) / \tau)}{\sum_{j=1}^{N} \exp(\text{cos}(F_i, V_{j+1}) / \tau)}
\end{equation}
here \( \tau \) is a learnable temperature parameter.

Different from ~\cite{george2024vital}, where they directly apply the CLIP loss on the time-aligned visuo-tactile images, we instead fuse the tactile observation with a masked current image to predict the future image. We make this choice for two main reasons. First, in ~\cite{george2024vital}, the tactile representation is conditioned on proprioceptive states, which are unavailable in our dataset before the success of SLAM. Second, since different tasks may have varying images but similar tactile observations, fusing a masked current image helps the network learn a more expressive tactile representation. Without sufficient masking, the alignment becomes trivial. 

After pre-training, we train a Diffusion Policy~\cite{chi2023diffusion} on the SLAM-filtered data. Following~\cite{chi2023diffusion}, we use a U-Net~\cite{ronneberger2015u} as the noise prediction network and apply DDIM~\cite{song2020denoising} to accelerate the inference for action prediction.


\section{Experiments}

\begin{figure}[htbp!]
    \centering
    \includegraphics[width=1.0\columnwidth]{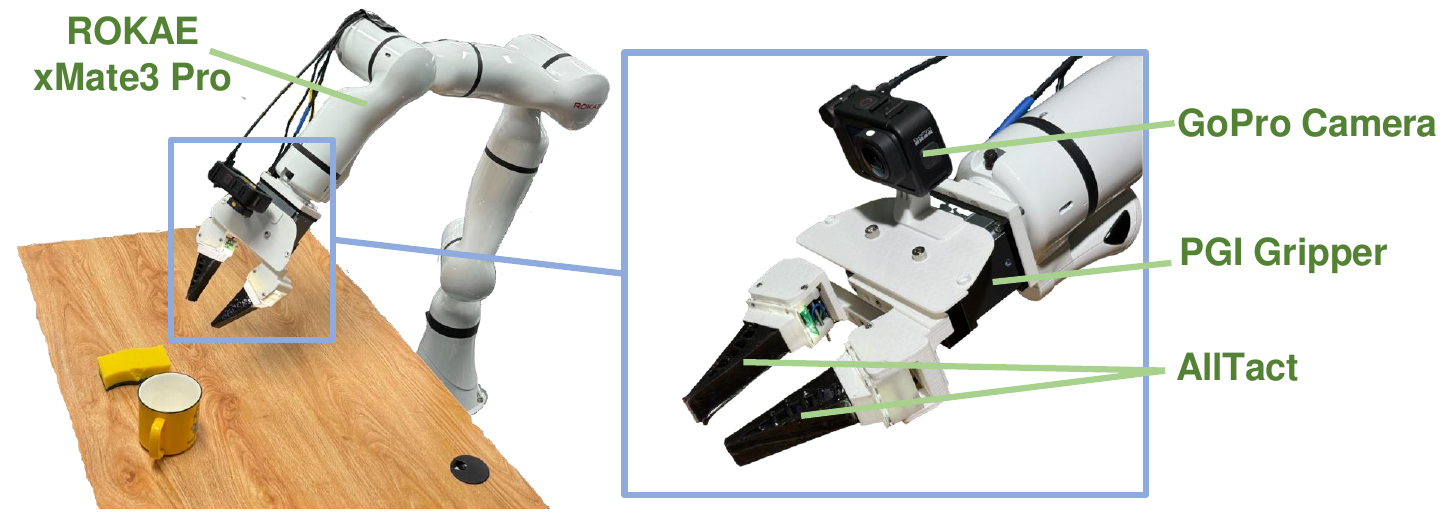}
    \caption{Hardware setup for policy deployment.}
    \label{fig:experiment-setup}
    \vspace{-3mm}
\end{figure}
\subsection{Experimental Setup}
\textbf{Hardware} 
Figure~\ref{fig:experiment-setup} shows the policy deployment setup. Our system consists of a Rokae xMate ER3PRO robotic arm equipped with a PGI-140-80-W-S parallel gripper. The 7-DOF robotic arm provides flexible manipulation capabilities, while the gripper features an 8cm stroke range from fully open to closed position. The system is implemented using ROS Noetic on Ubuntu 20.04. The control loop operates at 10Hz, with separate threads handling robot control, visual and tactile sensing. The system architecture is designed to minimize latency while maintaining reliable real-time performance.

Similar to UMI~\cite{chi2024universal}, our system compensates for various sources of latency in the perception-action loop through predictive buffering and timestamp-based synchronization between visual and tactile feedback streams. The policy generates 16 consecutive trajectories at each inference step, with 10 trajectories being executed based on our temporal compensation strategy.



\textbf{Manipulation Tasks}
\begin{figure*}[t]
\centering
\includegraphics[width=\textwidth]{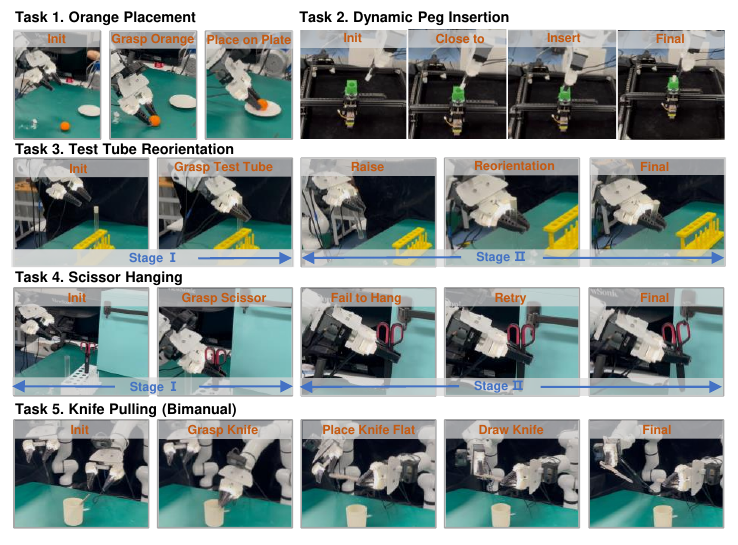}
\caption{We test \MethodAcronym{} on 5 contact-rich manipulation tasks, including precise and dynamic insertion, object hanging with multimodal feedback, and transparent in-hand object manipulation.}
\label{fig:tasks}
\vspace{-2mm}
\end{figure*}
As shown in Figure~\ref{fig:tasks}, we propose diverse contact-rich manipulation tasks to evaluate the effectiveness of \MethodAcronym{}. These tasks are specifically crafted to demonstrate the following key capabilities:
(1) \textit{Robust pick-and-place} of diverse objects, including fragile and small objects;
(2) \textit{Dexterous manipulation}, such as in-hand reorientation;
(3) \textit{Task success determination}, allowing the robot to repeat attempts until success;
(4) \textit{Dynamic and precise manipulation}.

We design the following 5 manipulation tasks:
\begin{itemize}
    \vspace{-2mm}
    \item Orange Placement: Put a fragile orange from a randomized position to a randomized plate.
    
    \item  Dynamic Peg Insertion: Grasp a peg and approach a hole, which is moving at a constant speed of 10 mm/s. And precisely insert the peg to the hole.
    
    \item  Test Tube Reorientation: Grasp a transparent test tube from a shelf and adjust its pose through extrinsic dexterity based on tactile feedback.
    
    \item  Scissor Hanging: Grasp a pair of scissors and hang them on a hook. Adjust the pose and keep attempting until it succeeds.
    
    \item Dual-Arm Knife Pulling: The left arm first grasps a knife from a cup, orients it horizontally. The right arm grasps and pulls it out with a constrained prismatic motion. This task requires tactile feedback to grasp the thin object and perform the correct pulling motion.
\end{itemize}

\label{sec:app_data}
\begin{table}[h!]
\centering
\caption{Data Collection Statistics for Different Tasks}
\label{tab:data_statistics}
\begin{tabular}{lccc}
\toprule
Task & Raw Data & Valid Data$^*$ & Avg. Length\\
\midrule
Orange Placement & 87 & 73 & 435 \\
Dynamic Peg Insertion & 201 & 141 & 321 \\
Test Tube Reorientation & 150 & 125 & 619 \\
Scissor Hanging & 172 & 137 & 642 \\
\midrule
Knife Pulling (Left) & 188 & 131 & 403 \\
Knife Pulling (Right) & 180 & 134 & 254 \\
\bottomrule
\multicolumn{4}{l}{\footnotesize $^*$Valid data refers to demonstrations with successful SLAM tracking} \\
\end{tabular}
\end{table}

Table~\ref{tab:data_statistics} shows the statistics of the demonstration data.
We collect demonstrations for both single-arm and dual-arm manipulation tasks. For single-arm tasks, we gather between 87 and 172 raw demonstrations per task according to the task difficulty, with successful SLAM tracking achieved in approximately 80\% of the trajectories. The dual-arm knife pulling task requires coordinated motion between both arms, with similar data collection volumes but slightly different average demonstration lengths for left and right arm movements.

We compare our approach against the following methods:
    (1) Vision: the policy only takes visual observation from the GoPro camera, which is encoded by the pre-trained CLIP model (identical to the original UMI~\cite{chi2024universal} paper); (2) Ours w/o Pre-training: This baseline simply concatenate visual and tactile observations after separate CLIP ViT-B/16 encoders, and fine-tuned with behavior cloning.

\begin{table}[htbp!]
    \centering
    \begin{tabular}{lccc}
    \toprule
    Task & Vision & w/o Pre-training & Ours  \\
    \midrule
    \multicolumn{4}{c}{\textit{Single-Arm Tasks}} \\
    \midrule
    Orange placement & 0.85 & 0.9 & \textbf{1} \\
    Test Tube Reorientation & 0.4 & 0.7 & \textbf{0.9} \\
    Scissor Hanging & 0.1 & 0.45  & \textbf{0.7} \\
    Dynamic Peg Insertion & 0.45 & 0.8  & \textbf{0.9} \\
    \midrule
    \multicolumn{4}{c}{\textit{Dual-Arm Task}} \\
    \midrule
    Knife Pulling & 0.6 & 0.8 & \textbf{0.9}  \\
    \bottomrule
    \end{tabular}
    \caption{Comparisons on 5 tasks with baselines. Our approach improves the performance on 5 tasks through multi-modal sensing and pre-training.}
    \label{tab:method_comparison}
    \vspace{-3mm}
\end{table}
The results are presented in Table~\ref{tab:method_comparison}. For each task, we conduct 20 trials with randomized initial conditions and report the average performance. The vision-only policy performs the worst across all five tasks, particularly in contact-rich tasks like test tube reorientation and scissor hanging, where tactile feedback is crucial for success. Across all tasks, pre-training enhances the performance, highlighting the importance of learning effective tactile representations.

\subsection{Failure Analysis}
In the Orange placement task, the robot picks up an orange from a random position within a 50\,cm$\times$50\,cm workspace and places it on a plate. Failures stem from table collisions, unstable placement, or motion planning errors despite correct object detection. In Dynamic peg insertion, the robot inserts a grasped peg into a moving hole. Vision-only methods often fail due to imprecise localization and alignment.

In Test tube reorientation, the robot must pick up a tube from a random rack location and reorient it vertically, with success defined by less than 10$^\circ$ orientation error. Failures include rack collisions, over-lifting, and incorrect final orientation. Scissor hanging requires picking up scissors and hanging them on a narrow hook, where common issues include misdetection, misalignment, and failure to release. 
In Knife pulling, a dual-arm policy reorients the knife with one arm while the other pulls it out of a holder. Failures often result from poor coordination, weak grasps, or incomplete pulling. Overall, vision-only policies struggle with contact-rich tasks, highlighting the limitations of unimodal sensing.

\subsection{Compliant Articulated Object Manipulation}
To demonstrate the compliance capabilities of ViTaMIn, we designed a compliant-controlled articulated object manipulation task. The robotic arm needs to grasp a handle (connected to a force gauge) and rotate it 90 degrees to open a switch. During the rotation process, the arm must minimize axial forces to ensure smooth operation.
We conduct 10 experiments for each condition and calculate the average forces. The results show that ViTaMIn achieves significantly lower average forces compared to using pure vision as input.

\begin{figure}[h]
    \centering
    \includegraphics[width=0.5\textwidth]{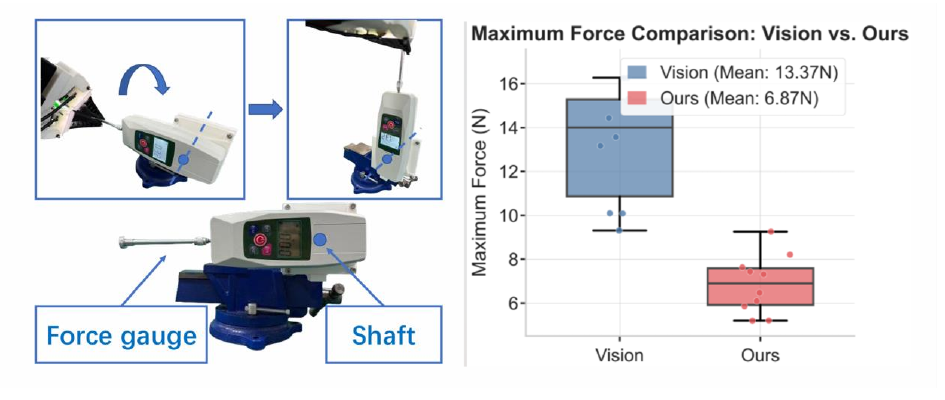}
    \caption{The robot needs to flip open a switch (fixed to a force gauge) by rotating it 90 degrees. During the rotation, the robot must minimize axial forces to ensure smooth operation.}
    \label{fig:test_force_task}
\end{figure}

\begin{figure}[h]
\includegraphics[width=1.0\linewidth]{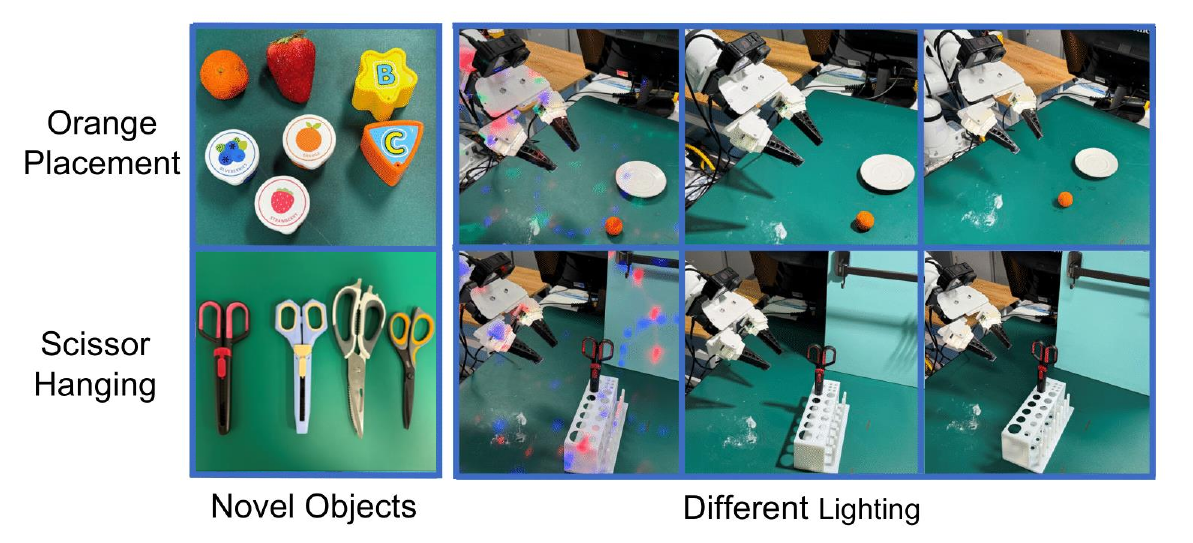}
    \caption{Showcase of novel objects and different lighting in the generalization tasks. The right columns demonstrates colored flashlight/high-power/normal lighting conditions.}
    \label{fig:generalizaton-setup}
\end{figure}

\subsection{Ablation Studies}
\begin{figure*}[!ht]
\centering
\includegraphics[width=\textwidth]{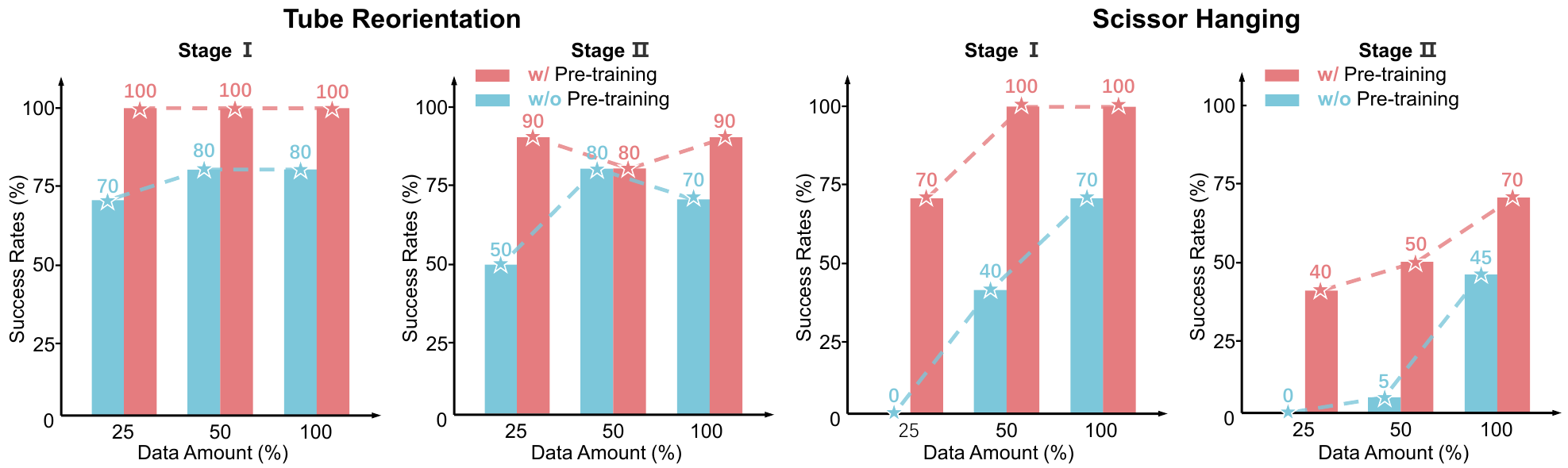}
\caption{Ablation study on the effect of pre-training on data efficiency. The performance of the policy improves as the quantity of data increases. After pre-training on the action-free, task-ignorant dataset, our method can achieve a high success rate even with limited data (25\%).}
\label{fig:ablation_data_efficiency}
\vspace{-2mm}
\end{figure*}

\paragraph{Data Efficiency}
We evaluate the performance of policies trained on different amounts (25\%, 50\%, and 100\%) of demonstrations. All the models are evaluated in 20 real-world trials with different initializations. For a more in-depth analysis, we calculate the success rates of each stage separately, as illustrated in Figure~\ref{fig:ablation_data_efficiency}. With the pre-trained tactile representations, our method can achieve consistently higher success rates on all the tasks across different amounts of data, and can even master the task with limited data (25\%) for test tube reorientation.

\begin{figure*}[!htbp]
\centering
\includegraphics[width=\textwidth]{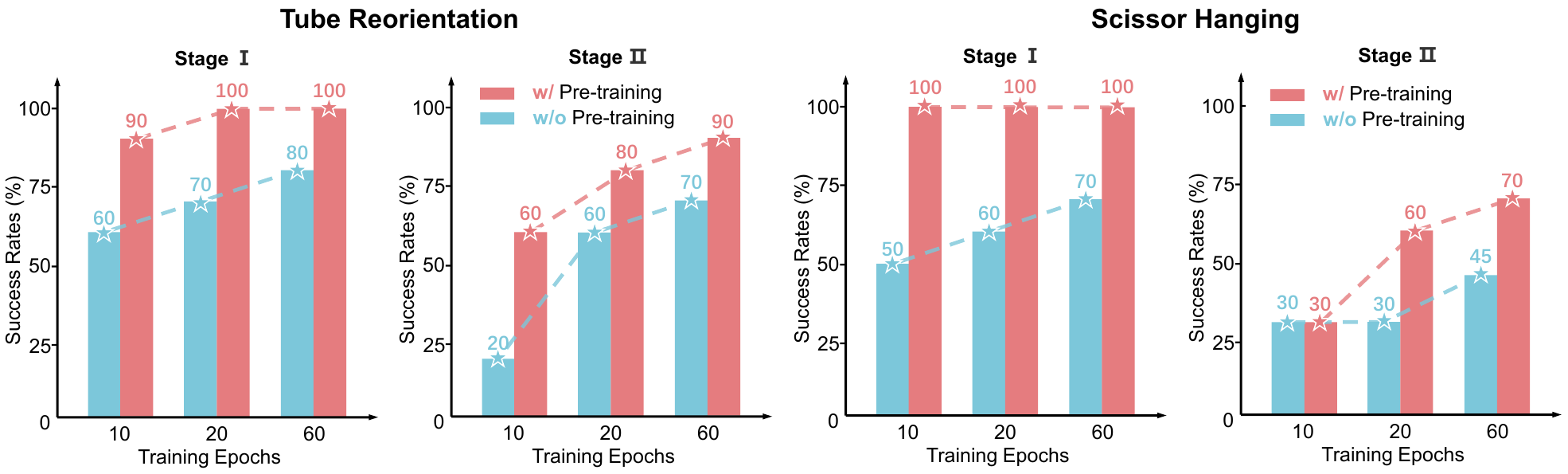}
\caption{Ablation study on the effect of pre-training on training efficiency. Policies with pre-training are able to learn to complete the first-stage task at a remarkably early stage of training (within 10 epochs). Additionally, when the policy network is pre-trained, the overall success rates increase more rapidly.}
\label{fig:ablation_training_efficiency}
\vspace{-3mm}
\end{figure*}

\paragraph{Training Efficiency}
We further evaluate the policies trained with different numbers of epochs to understand its training efficiency under the same evaluation protocol. The results are illustrated in Figure~\ref{fig:ablation_training_efficiency}. We also observe consistent task performance improvements with pre-training. The policy can complete the first stage of the task at a remarkably early training stage (within 10 epochs).

\begin{table}[htbp!]
    \centering
    \begin{tabular}{c|l|ccc}
    \toprule
    Task & Method & Original & \makecell{ Novel 
 \\Objects }& \makecell{ Different\\ Lighting} \\
    \midrule
     \multirow{3}{*}{\makecell{Orange\\ Placement}} & Vision & 0.85 & 0.7 & 0.55 \\
     & Ours w/o Pre-training & 0.9 & 0.8 & 0.6 \\
      & Ours & \textbf{1.0} & \textbf{1.0} & \textbf{0.85} \\
    \midrule
     \multirow{3}{*}{\makecell{Scissor \\ Hanging}}& Vision & 0.0 & 0.0 & 0.0 \\
     & Ours w/o Pre-training & 0.45 & 0.4 & 0.4 \\
      & Ours & \textbf{0.7} & \textbf{0.7} & \textbf{0.5} \\
    \bottomrule
    \end{tabular}
    \caption{Generalization under different objects and scenes. The results demonstrate that our multi-modal policy is more robust to novel objects and different lighting conditions.}
    \label{tab:generalization}
\end{table}

\subsection{Generalization Capability}
We also evaluate our policy’s generalizability to unseen objects and environments. As shown in Figure~\ref{fig:generalizaton-setup}, beyond the training orange and scissor, we introduce 6 unseen small objects and 3 unseen scissors to assess object generalization. Additionally, we modify lighting conditions by increasing brightness and introducing colored disco ball lighting. 
Table~\ref{tab:generalization} presents results on the tasks of orange placement and scissor hanging. Our method with pre-training achieves consistent better performance across various generalization settings.

\section{Conclusion}
In this paper, we present \MethodAcronym{}, a portable visuo-tactile manipulation interface designed for efficiently collecting high-quality demonstrations by capturing both visual and tactile signals. Furthermore, \MethodAcronym{} introduces an effective pre-training strategy that leverages all the collected action-free data to learn a robust and generalizable tactile representation through multimodal contrastive learning. Our approach significantly outperforms vision-only policies across 5 real-world contact-rich manipulation tasks and demonstrates improved data efficiency, robustness, and generalizability with pre-trained visuo-tactile representations.

Our method primarily focuses on fixed-base single-arm and dual-arm tasks with parallel-jaw grippers. While this setup is suitable for a wide range of manipulation tasks, future work could extend our approach to dexterous hands, enabling richer and more versatile manipulation skills that better approximate human-level dexterity.

{%
\bibliographystyle{IEEEtran}
\bibliography{ref}
}

\clearpage

\end{document}